\documentclass[lettersize,journal]{IEEEtran}
\usepackage[utf8]{inputenc}
\usepackage{soul}
\usepackage{lettrine}
\usepackage{cite}
\usepackage{supertabular,booktabs}
\usepackage{float}
\usepackage{comment}
\usepackage{makecell}
\usepackage{graphicx}
\usepackage{array}
\usepackage{amsfonts}
\usepackage{pdfpages}
\usepackage{amsmath}
\usepackage{algorithm}
\usepackage{algpseudocode}
\usepackage{xcolor}

\usepackage{subfig}
\usepackage{amssymb}
\usepackage{relsize}

\usepackage[hyphens]{url}
\usepackage{xurl}

\usepackage{hyperref}
\hypersetup{hidelinks}

\title{FL-PBM: Pre-Training Backdoor Mitigation for Federated Learning}

\author{
    Osama Wehbi, Sarhad Arisdakessian, Omar Abdel Wahab, Azzam Mourad, Hadi Otrok, Jamal Bentahar\\
    \thanks{
        Osama Wehbi is with the Department of Computer and Software Engineering, Polytechnique Montréal, Montreal, Quebec, Canada (e-mail: osama.wehbi@etud.polymtl.ca).

        Sarhad Arisdakessian is with the Department of Computer and Software Engineering, Polytechnique Montréal, Montreal, Quebec, Canada (e-mails: sarhad.arisdakessian@etud.polymtl.ca).

        Omar Abdel Wahab is with the Department of Computer and Software Engineering, Polytechnique Montréal, Montreal, Quebec, Canada (e-mails: omar.abdul-wahab@polymtl.ca).

        Azzam Mourad is with the Department of Computer Science, Khalifa University, Abu Dhabi, UAE, as well as the Artificial Intelligence \& Cyber Systems Research Center, Department of CSM, Lebanese American University (e-mails: azzam.mourad@ku.ac.ae).
        
        Hadi Otrok is with the Department of Computer Science, Khalifa University, Abu Dhabi, UAE (e-mails: hadi.otrok@ku.ac.ae).
        
        Jamal Bentahar is with the KU 6G Research Center, Department of Computer Science, Khalifa University, Abu Dhabi, UAE, and also with the
        Concordia Institute for Information Systems Engineering, Concordia University, Montreal, Quebec, Canada (e-mail: jamal.bentahar@ku.ac.ae).

    }
}
\begin{document}

\maketitle

\begin{abstract}

Backdoor attacks pose a significant threat to the integrity and reliability of Artificial Intelligence (AI) models, enabling adversaries to manipulate model behavior by injecting poisoned data with hidden triggers. These attacks can lead to severe consequences, especially in critical applications such as autonomous driving, healthcare, and finance. Detecting and mitigating backdoor attacks is crucial across the lifespan of model's phases, including pre-training, in-training, and post-training. In this paper, we propose Pre-Training Backdoor Mitigation for Federated Learning (FL-PBM), a novel defense mechanism that proactively filters poisoned data on the client side before model training in a federated learning (FL) environment. The approach consists of three stages: (1) inserting a benign trigger into the data to establish a controlled baseline, (2) applying Principal Component Analysis (PCA) to extract discriminative features and assess the separability of the data, (3) performing Gaussian Mixture Model (GMM) clustering to identify potentially malicious data samples based on their distribution in the PCA-transformed space, and (4) applying a targeted blurring technique to disrupt potential backdoor triggers. Together, these steps ensure that suspicious data is detected early and sanitized effectively, thereby minimizing the influence of backdoor triggers on the global model. Experimental evaluations on image-based datasets demonstrate that FL-PBM reduces attack success rates by up to 95\% compared to baseline federated learning (FedAvg) and by 30 to 80\% relative to state-of-the-art defenses (RDFL and LPSF). At the same time, it maintains over 90\% clean model accuracy in most experiments, achieving better mitigation without degrading model performance.

\end{abstract}

\begin{IEEEkeywords}
Federated Learning (FL), Backdoor Attacks, PCA, GMM, Blurring.
\end{IEEEkeywords}

\section{Introduction}
\label{intro}

Artificial Intelligence has seen significant growth and adoption across various domains, including healthcare, autonomous driving, finance, and cybersecurity. The increasing reliance on AI-driven systems has highlighted the critical need for developing reliable and secure AI models that can effectively utilize the vast and diverse data generated at the edge while ensuring data privacy. Leveraging edge data enables AI systems to learn from diverse and distributed environments, enhancing their adaptability and robustness. However, this distributed nature also opens up vulnerabilities to adversarial threats like backdoor attacks, which can compromise model integrity and lead to severe consequences, especially in safety-critical applications.

Federated Learning (FL) has emerged as a promising solution for collaborative model training while preserving client data privacy. Unlike centralized learning, FL allows clients to train models locally and share only the model updates with a central server, minimizing the risk of data leakage. This approach not only respects data privacy but also capitalizes on the diversity of edge devices resources and local data, making AI models more representative and effective. Recent research has shown that pre-processing data on the client side before training can be effective without compromising privacy, making it a viable strategy to mitigate adversarial attacks before data reaches the training phase~\cite{timofte2025federated}.

Backdoor attacks pose a significant threat to FL, as adversaries can inject poisoned data with hidden triggers. In this context, a trigger refers to a small, intentionally crafted pattern (e.g., a patch, pixel pattern, or feature perturbation) that forces the model to output an adversary-chosen label during inference.  Mitigation strategies for backdoor attacks generally focus on three phases, namely (1) Post-training, (2) In-training, and (3) Pre-training defenses. Post-training methods often employ techniques such as model pruning, neuron clustering, or activation pattern analysis to identify and remove backdoors after the global model has been trained. In-training approaches typically involve anomaly detection, robust aggregation rules, or gradient filtering to limit the influence of malicious updates during the learning process~\cite{nguyen2024backdoor}. While a substantial body of research has focused on post-training and in-training strategies, pre-training defenses particularly those applied at the client side through data cleaning, normalization, or trigger sanitization remain comparatively underexplored~\cite{li2025threats}.

In this work, we introduce \textit{Pre-Training Backdoor Mitigation for Federated Learning (FL-PBM)}, a proactive client-side defense mechanism designed to operate within a Trusted Execution Environment (TEE), ensuring that the mitigation process is isolated and cannot be altered~\cite{TEEsref}. \textit{FL-PBM} identifies and neutralizes poisoned data before they can compromise the global model. To enable efficient separation between clean and poisoned samples in high-dimensional feature spaces, \textit{FL-PBM} employs Principal Component Analysis (PCA) for feature extraction followed by Gaussian Mixture Model (GMM) clustering. Based on~\cite{tran2018spectral}, backdoored samples exhibit distinct spectral characteristics in feature space, making PCA and GMM suitable for capturing these statistical deviations compared to alternative fixed cluster shapes techniques. Since our evaluation focuses on image-based datasets, data deemed highly suspicious are excluded from training, while moderately suspicious samples undergo targeted image transformations, such as blurring, to diminish the effect of potential backdoor triggers.

\subsection{Problem Statement}
\label{psm}
Backdoor attacks in FL introduce hidden vulnerabilities into AI models by embedding poisoned data samples containing covert triggers that, when activated, can manipulate predictions toward an attacker’s goal. These threats are especially critical in domains such as autonomous driving, healthcare, and finance, where malicious outputs can cause severe or even life-threatening consequences. The decentralized and privacy-preserving design of FL gives adversaries opportunities to inject backdoors during client-side training without easily being detected by the central server. By remaining dormant during normal operation and activating only under specific input conditions, these attacks often bypass conventional performance metrics such as accuracy, making them challenging to detect.

Existing defense approaches are largely concentrated in two areas: post-training detection, which inspects the model after aggregation, and in-training defenses, which filter or control the model weights of suspicious updates during training. While these techniques have shown effectiveness, they tend to overlook early contamination of the training process, leaving the global model influenced by poisoned data from the very first rounds. In addition, many of these methods are tailored to narrow threat models such as single-trigger or targeted attacks and struggle against more complex scenarios like multi-trigger or collusion-based strategies.

The absence of robust pre-training defenses intensifies the vulnerability of FL systems to backdoor attacks, as poisoned data can enter the training process unchecked and compromise the global model. A truly effective solution must act before poisoned data affects the model, adapt to evolving client behaviors, and remain effective across diverse attack strategies. To meet these needs, in this work we propose Adaptive Pre-Training Backdoor Attack Mitigation for Federated Learning (FL-PBM), that proactively mitigates backdoor threats at the source, preventing poisoned data from compromising the training process and thereby preserving both model integrity and client data privacy before even the training starts.

\subsection{Contributions}
\label{contributions}

Motivated by the need for proactive defenses in FL environments as highlighted in \ref{psm}, this work proposes a novel pre-training backdoor mitigation strategy, called \textit{FL-PBM}. Unlike traditional defenses that focus on in-training or post-training mitigation, \textit{FL-PBM} operates in the early stages of training to filter poisoned data before it affects the global model. By leveraging distributed detection and transformation techniques, our approach significantly reduces the success rate of backdoor attacks while preserving model performance. 

Accordingly, the main contributions of our work can be summarized as follows:

\begin{enumerate}
    \item To the best of our knowledge, we are from the firsts to introduce a client-side pre-training backdoor mitigation technique in FL, proactively identifying and neutralizing poisoned data before local models are trained and aggregated.
    \item Combining PCA-based feature extraction with GMM clustering and benign trigger insertion to detect suspicious samples in an unsupervised manner, addressing the challenge of distinguishing subtle feature deviations of backdoored data in high-dimensional spaces.
    \item Applying selective data-driven blurring at the client-side prior to training to suppress the effectiveness of hidden triggers, thus weakening the influence of injected backdoors without degrading overall accuracy.
    \item Designing our solution in such a way that it remains effective against a wide spectrum of backdoor attack strategies, including one-to-one and multi-trigger scenarios, thereby ensuring broad applicability across real-world federated learning deployments.
\end{enumerate}

These contributions advance the state-of-the-art in proactive backdoor mitigation for FL by addressing threats at the earliest stage of training. FL-PBM’s privacy-preserving and attack-agnostic design enables practical deployment in diverse, real-world FL applications, paving the way for more secure and trustworthy collaborative learning systems. Empirical evaluations demonstrate that \textit{FL-PBM} substantially mitigates the attack, reducing success rates by more than 95\% relative to baseline methods. Competing approaches such as RDFL and LPSF show notably lower resistance, curbing attacks by roughly 20-70\% and 10-60\%, respectively. Importantly, \textit{FL-PBM} sustains clean model accuracy above 90\% in both IID and non-IID scenarios across most experiments.

\subsection{Paper Outline}
The rest of the this paper is organized as follows: Section~\ref{related work} reviews the related work on backdoor attack detection and mitigation strategies in FL. Section~\ref{system preliminaries} discusses the underlying concepts that drive this work, and highlight the methodology used. Section~\ref{problem formulation} formulates the backdoor attack problem in FL and the corresponding defense mechanisms. We provide the mathematical formulations of the process including PCA projections, GMM-based anomaly detection, adaptive blurring, as well as the adversarial objective, which together establish the basis of our defense framework. Section~\ref{FL-PBM} describes the implementation of \textit{FL-PBM} and its operation across multiple training rounds. Section~\ref{exp} reports on the experimental results, evaluating \textit{FL-PBM} effectiveness using real-world datasets and comparing its performance with state-of-the-art methods in terms of model accuracy and backdoor attack success rates. Section~\ref{conc} concludes the paper, summarizing the key contributions and suggesting potential directions for future research.

\section{Related Work}
\label{related work}

In this section, we review the state-of-the-art pertaining to backdoor attacks in FL, highlighting their strengths and limitations to motivate our proposed approach.

The authors of~\cite{lin2024mitigating} propose Federated Layer Detection (FLD), a pre-aggregation method that analyzes model updates at the layer level to detect backdoor attacks in federated learning (FL). By examining layers individually, FLD captures fine-grained differences between benign and malicious updates, making it effective against complex attacks. However, the approach can introduce computational overhead and struggles with varying model architectures and adaptive attackers who mimic benign patterns. In~\cite{GU2025125359}, the authors present ANODYNE, which uses perturbation techniques like differential privacy and weight clipping to suppress anomalous updates. While enhancing robustness by injecting noise, the method risks degrading model accuracy and increasing computational costs, focusing on sanitization before training. In~\cite{gill2023feddefender}, the authors propose FedDefender, which employs statistical anomaly detection during training to filter suspicious client updates, effectively reducing backdoor impact. Nevertheless, it incurs high computational cost and may not be effective against attackers who closely imitate legitimate updates. The authors of ~\cite{10992396} propose Robust Knowledge Distillation, which mitigates backdoors by transferring distilled knowledge instead of raw updates, addressing non-IID data challenges and high dimensionality. Yet, reliance on a teacher model introduces bias and computational overhead, while operating during post-training. The game-theoretic proposed approach called  FedGame in~\cite{jia2023fedgame} anticipates adversarial behaviors to detect backdoors proactively during training by analyzing client interactions. Despite its proactive stance, it depends on predefined attack models, limiting its effectiveness against novel threats. On the other hand, the adaptive attack study in~\cite{zhang2023a3fl} exposes vulnerabilities in FL by optimizing trigger patterns based on training dynamics, underscoring challenges existing defenses face against evolving backdoors. 

In~\cite{huang2025scope}, the authors propose an approach called SCOPE that targets constrained backdoor attacks by applying statistical and anomaly detection techniques during aggregation. While effective against sophisticated attacks, its performance may degrade with highly heterogeneous client data and inconsistent benign behaviors, risking false positives or negatives. In~\cite{zhang2025runtime}, the authors propose Runtime defense via Representational Dissimilarity Analysis (RDA), which iteratively excludes anomalous client updates during aggregation, enhancing robustness with low overhead. However, The approach might struggle when malicious updates closely resemble benign ones or in environments with diverse client data. In~\cite{chen2025mdsd}, the authors propose MDSD, that combines multi-dimensional scaling with norm clipping and weighted aggregation to identify and reduce malicious contributions. Though effective in lowering attack success, it may be less reliable when attackers mimic benign updates in heterogeneous settings. Meanwhile, the dual-layer defense in fog-based FL by~\cite{gu2024dual} uses gradient metrics and adaptive filtering at fog and central servers to detect and mitigate attacks, but may face challenges with unseen attack types and highly non-IID data. 

The work in~\cite{zhang2024flpurifier} proposes FLPurifier, which employs decoupled contrastive training to weaken trigger-target associations and adaptively aggregates classifiers, maintaining accuracy while defending backdoors. However, it is limited to models where feature-classifier decoupling is feasible. The authors in~\cite{zhang2023safelearning} propose SAFELearning, which highlights secure aggregation with oblivious random grouping and partial parameter disclosure to detect backdoors without compromising privacy. Despite its effectiveness, secure multi-party computations add computational overhead. The comprehensive framework in~\cite{lu2022defense} combines client-side data sanitization and server-side robust aggregation, lowering attack success but facing difficulties with non-IID data and adaptive attackers. The authors of~\cite{rodriguez2022backdoor} propose robust aggregation that filters outlier updates statistically to defend backdoors in image classification. However, effectiveness declines when malicious updates resemble benign ones or in highly heterogeneous data scenarios. 

On the other hand, FedBlock~\cite{nguyen2024fedblock} leverages blockchain to secure FL by ensuring transparency and traceability of client updates, improving defense against backdoors. Yet, blockchain overhead may limit scalability in large networks. The authors of~\cite{kaviani2023defense} propose LPSF, which introduces Link-Pruning with Scale-Freeness that strengthens essential connections and prunes dormant links in feed-forward neural networks to enhance robustness against backdoor attacks. While it achieves 50-94\% reduction in attack success on several datasets, its evaluation is limited to feed-forward networks and may not generalize to deeper architectures. Finally, the work in ~\cite{wang2023adaptive} proposes, a technique called RDFL; an adaptive FL defense approach that relies on cosine-distance-based parameter selection, adaptive clustering, malicious model removal, and clipping with noise. It effectively mitigates backdoor attacks in non-IID scenarios.However, the approach might incur higher computation and communication overhead in large-scale federated networks.

While existing defenses against backdoor attacks in FL have provided valuable insights, most approaches rely on in-training or post-training mitigation, centralized verification, and static assumptions about attacker behavior, which limit their adaptability in dynamic environments. Moreover, many methods overlook early-stage client-side defenses that could intercept poisoned data before it propagates into the global model. In contrast, our proposed \textit{FL-PBM} adopts a proactive, pre-training defense strategy that intervenes at the very beginning of the learning process. By integrating Principal Component Analysis (PCA) for discriminative feature extraction, Gaussian Mixture Model (GMM) clustering for anomaly detection, and an adaptive blurring mechanism to sanitize suspicious samples, \textit{FL-PBM} mitigates poisoned data before it influences model aggregation; disrupting backdoor threats at their origin while preserving model integrity and client privacy.

\section{System Preliminaries}
\label{system preliminaries}
In this section, we introduce the fundamental concepts that underpin our work. We begin with a high-level overview of the FL paradigm, then describe backdoor attack strategies, and finally detail the techniques we employ for pre-training backdoor mitigation, including PCA, GMM, and adaptive blurring.

A high-level overview of the FL paradigm is provided, followed by a description of backdoor attack strategies, and finally the techniques employed for pre-training backdoor mitigation, including PCA, GMM, and adaptive blurring, are detailed.

\subsection{Federated Learning Process}
Federated Learning is a distributed machine learning framework that enables multiple clients (e.g., autonomous vehicles, edge nodes, or organizations) to collaboratively train a shared global model under the coordination of a central server, without exchanging their raw data. Instead, clients perform local training on their private datasets and periodically communicate model updates to the server. The server then aggregates these updates to improve the global model, which is subsequently redistributed to clients for the next training round.

The general workflow of FL can be summarized as follows:
\begin{enumerate}
    \item \textbf{Model Initialization:} The server initializes a global model and shares it with selected clients.
    \item \textbf{Local Training:} Each client trains the received model on its private dataset, generating updated local parameters.
    \item \textbf{Aggregation:} The server collects local model updates and combines them into a new global model.
    \item \textbf{Iteration:} Steps 2 and 3 are repeated over multiple communication rounds until the model converges to a desired accuracy, or the process stops after a certain amount of rounds.
\end{enumerate}

\subsection{Backdoor Attacks in Federated Learning}
Backdoor attacks are a type of targeted poisoning attack in which adversaries deliberately manipulate local training data or model updates to implant hidden behaviors in the global model. A successful backdoor attack ensures that the model performs normally on clean data but produces attacker-controlled misclassifications when a specific trigger pattern is present in the input.

In FL, backdoor attacks are often executed by malicious clients that introduce poisoned samples into their local datasets. These poisoned samples typically contain a visual trigger (such as a small patch or specific pixel pattern) and are mislabeled to a target class. Over multiple training rounds, the global model learns to associate the trigger with the target label, enabling the attacker to control its predictions at inference time.

Backdoor attacks can be grouped as follows:
\begin{itemize}
    \item \textbf{One-to-One (1$\rightarrow$1):} A single trigger causes misclassification into a single target class.
    \item \textbf{One-to-N (1$\rightarrow$N):} A single trigger with different variations each mapped to distinct target classes, improving attack robustness.
    \item \textbf{N-to-One (N$\rightarrow$1):} Multiple different triggers are mapped to the same target class, often corresponding to multiple malicious participants.
\end{itemize}

The privacy-preserving and decentralized nature of FL makes it particularly challenging to detect such attacks, as the server lacks direct access to client training data.

\subsection{Pre-Training Backdoor Mitigation Techniques}
To address the threat of backdoor attacks, in this work, we focus on client-side pre-training defenses. These defenses aim to identify and neutralize suspicious samples before they influence local model updates. Our strategy relies on unsupervised methods to analyze data distributions and highlight anomalies, while preserving privacy by keeping all computations local to each client.

\subsubsection{Principal Component Analysis (PCA)}
Principal Component Analysis (PCA) is an unsupervised dimensionality reduction technique that transforms the data into a new coordinate system, where the axes (principal components) correspond to directions of maximum variance in the data~\cite{jolliffe2016principal}. Unlike supervised methods, PCA does not rely on class label information, making it more suitable in our context, where labels may be unreliable due to backdoor attacks. By projecting data onto the principal components, PCA captures the dominant patterns of variation in the input space. This helps highlight deviations introduced by poisoned samples, as they tend to disturb the natural variance structure of clean data. Consequently, PCA provides a robust feature representation that improves anomaly detection accuracy in the presence of mislabeled or manipulated data.

\subsubsection{Gaussian Mixture Model (GMM) Clustering}
The Gaussian Mixture Model (GMM) is a probabilistic clustering technique that models data as a combination of multiple Gaussian distributions~\cite{campos2025federated}. After PCA projection, GMM is applied to capture the underlying structure of the transformed data. Samples that fall into low-likelihood clusters or significantly deviate from cluster centers are flagged as suspicious. This unsupervised detection complements PCA by modelling hidden data patterns relying on distributional characteristics rather than label information, making it particularly effective in spotting backdoor triggers that subtly distort data distributions.
 
PCA emphasizes the dominant variance structure in the data without relying on labels, while GMM models probabilistic density and captures outliers.

\subsection{Adaptive Image Blurring Mechanism}
Image blurring is a common image processing operation that reduces high-frequency details by smoothing pixel intensity variations, often through convolution with a low-pass filter such as a Gaussian kernel \cite{shi2023black}. By attenuating fine-grained structures, blurring suppresses subtle patterns while preserving the overall semantic content of an image.

In our framework, once suspicious samples are identified through the PCA-GMM pipeline, we apply an adaptive blurring strategy as a mitigation step. The approach is two-fold:
\begin{itemize}
    \item \textbf{High-risk samples:} Completely excluded from training to avoid contamination.
    \item \textbf{Moderate-risk samples:} Processed using adaptive blurring, which reduces fine-grained pixel details where triggers are likely embedded, while retaining the semantic features necessary for effective learning.
\end{itemize}

This selective mechanism balances data utility and security. It degrades malicious triggers without discarding all suspect data. The adaptive blurring reduces the attacker’s effectiveness while preserving training diversity and accuracy.

\section{Problem Formulation}
\label{problem formulation}

In this section, we provide the mathematical formulation of our problem setting. We first formalize the FL framework, then describe the backdoor attack model, and finally formulate the pre-training mitigation components, namely PCA, GMM, and adaptive blurring. Additionally, Table~\ref{tab:symbols} summarizes the key notations used throughout this section.

\begin{table}[ht]
\centering
\begin{tabular}{|c|p{5.5cm}|}
\hline
\textbf{Symbol} & \textbf{Description} \\
\hline
$\mathcal{N}$ & Number of participating clients \\
$\mathcal{D}_i$ & Local dataset of client $i$ \\
$\mathcal{G}^t$ & Global model at communication round $t$ \\
$\Lambda_i^t$ & Local model of client $i$ at round $t$ \\
$L(\cdot)$ & Local loss function \\
$\eta$ & Local learning rate \\
$x, y$ & Input sample and label pair \\
$x^{\prime}$ & Poisoned (triggered) input sample \\
$\tau$ & Trigger pattern for backdoor attack \\
$y^a$ & Target class assigned by attacker \\
$\phi$ & PCA projection matrix\\
$z$ & Data representation after PCA projection \\
$\theta_y = \{w, \mu, \Sigma\}$ & GMM parameters for label $y$ (weights, means, covariances) \\
$p(z|\theta)$ & Likelihood of data under GMM \\
$P_B(\omega)$ & Mean power spectral densities (PSDs) of benign set\\
$d_p$ & Distance of image $p$ to malicious cluster center \\
$\delta_p$ & Normalized closeness of suspicious image $p$ \\
$\sigma_p$ & Gaussian blur strength for image $p$ \\
$G_{\sigma_p}$ & Gaussian kernel with standard deviation $\sigma_p$ \\
$I'_p$ & Defended (blurred) version of image $p$ \\
\hline
\end{tabular}
\caption{List of key symbols used in problem formulation.}
\label{tab:symbols}
\end{table}

\subsection{Federated Learning Framework}
We consider a federated learning system comprising $\mathcal{N}$ clients, where each client $i$ possesses a private dataset $\mathcal{D}_i = \{(x_j, y_j)\}_{j=1}^{|\mathcal{D}_i|}$, where $x_{j}$ denotes the $j$-th input sample of client $i$ and $y_{j}$ is its corresponding label. The global model $\mathcal{G}^t$ is trained collaboratively under the coordination of a central server. At each communication round $t$, the server broadcasts the current global model $\mathcal{G}^{t-1}$ to a subset of clients. Each selected client performs local loss optimization

\begin{equation}
\Lambda_i^t \in \arg\min_{\Lambda_i^t} \sum_{(x,y)\in \mathcal{D}_i} L(\Lambda_i^t, x, y)
\end{equation}
and returns the updated local model $\Lambda_i^t$ to the server. The server aggregates the local updates via a weighted average:
\begin{equation}
\label{eq:agg}
\mathcal{G}^t = \sum_{i=1}^{\mathcal{N}} \frac{|\mathcal{D}_i|}{\sum_{j=1}^{\mathcal{N}}|\mathcal{D}_j|} \, \Lambda_i^t
\end{equation}

\subsection{Backdoor Attack Formulation}

Backdoor attacks aim to implant a hidden mapping in the model that is only triggered under specific conditions. These attacks are typically realized by modifying the training data at malicious clients. The attacker introduces a trigger $\tau$ and changes the label to a target class $y^a$. We classify the attacks into three types:

\subsubsection{One-to-One Attack}
A single static trigger $\tau$ is applied uniformly, leading to a fixed target label:
\begin{equation}
\label{eq:back1}
\mathcal{D}_i' = \{ (x_j + \tau, y_j^a) : (x_j, y_j) \in \mathcal{D}_i \}
\end{equation}
This ensures that whenever the trigger $\tau$ is present, the model outputs $y_j^a$, regardless of the original class.

\subsubsection{One-to-N Attack}
A trigger family $\{\tau_k\}$ is used, with different variants $\tau_k$ causing misclassification into multiple target labels:
\begin{equation}
\label{eq:back2}
\mathcal{D}_i' = \{ (x_j + \tau_k, y_j^a) : (x_j, y_j) \in \mathcal{D}_i,\ k = 1, \ldots, K \}
\end{equation}
where $k$ is the total number of trigger variants, each variation $\tau_k$ may correspond to a different adversarial target, increasing the attack’s complexity and evasiveness.

\subsubsection{N-to-One Attack}
Multiple clients inject distinct triggers $\tau_k$ designed to collectively activate the same target behavior when combined:
\begin{equation}
\label{eq:back3}
\mathcal{D}_i' = \{ (x_j + \sum_{k=1}^{\mathcal{N}} \tau_k, y_j^a) : (x_j, y_j) \in \mathcal{D}_i \}
\end{equation}
The attack only activates when all components $\tau_k$ are present together, making it extremely difficult to detect when viewed in isolation.

The adversary’s objective is to minimize the global loss on benign data while simultaneously maximizing the misclassification rate of triggered inputs:
\begin{equation}
\min_{\mathcal{G}} \Bigg( \mathbb{E}_{(x,y)\sim \mathcal{D}_{\text{clean}}}[L(\mathcal{G},x,y)] 
+ \lambda \, \mathbb{E}_{(x^{\prime},c_t)\sim \mathcal{D}_{\text{poison}}}[L(\mathcal{G},x^{\prime},c_t)] \Bigg)
\end{equation}
where $\mathbb{E}$ stands for the expectation and $\lambda$ balances clean accuracy and attack success.

\subsection{Principal Component Analysis (PCA)}
For each client’s dataset $\mathcal{D}_i$, we perform PCA to obtain a projection matrix $\phi$ that captures the principal components of variance~\cite{jolliffe2011principal}. Specifically, we compute the covariance matrix of the data:
\begin{equation}
\label{eq:pca_cov}
\Sigma = \frac{1}{|\mathcal{D}_i|} \sum_{x \in \mathcal{D}_i} (x - \mu)(x - \mu)^\top,
\end{equation}
where $\mu$ is the mean of all samples in $\mathcal{D}_i$. PCA then solves the eigenvalue decomposition:
\begin{equation}
\Sigma \phi = \lambda \phi,
\end{equation}
where $\lambda$ and $\phi$ are the eigenvalues and eigenvectors of $\Sigma$. The eigenvectors corresponding to the top-$k$ largest eigenvalues form the projection matrix $\phi_k$.

The data is then projected into the reduced space as:
\begin{equation}
\label{eq:pca}
z = \phi_k^\top x
\end{equation}
where $z$ represents the lower-dimensional feature representation.

By retaining the directions of maximum variance, PCA emphasizes the dominant structure in the data while filtering out noise. This unsupervised projection is particularly robust in the presence of mislabeled or backdoored samples, since it does not rely on class labels.

\subsection{Gaussian Mixture Model (GMM)}
After PCA projection for class $y$, a two-component GMM is applied to the transformed data~\cite{zong2018deep}:
\begin{equation}
\label{eq:fit}
p(z \mid \theta_y) = \sum_{q=1}^{2} w_q \, \mathcal{N}(z \mid \mu_q, \Sigma_q)
\end{equation}
where $q \in \{1, 2\}$ indexes the Gaussian components, with each sample assigned to the Maximum A Posteriori (MAP) component:
\begin{equation}
\label{eq:assi}
\hat{q}(z) = \arg\max_{q \in \{1,2\}} w_q \, \mathcal{N}(z \mid \mu_q, \Sigma_q)
\end{equation}
The cluster containing the largest number of samples is designated as the benign set $C_{B,y}$, while the remaining cluster constitutes the suspicious set $C_{S,y}$:
\begin{equation}
\label{eq:cluster}
C_{B,y} = \arg\max_{C \in \{C_1, C_2\}} |C|, \qquad C_{S,y} = \{C_1, C_2\} \setminus C_{B,y}
\end{equation}

\subsection{Cluster Validation using Silhouette Score}
To assess the quality of clustering, we compute the \emph{Silhouette Coefficient}~\cite{shahapure2020cluster} using the PCA-projected features $z$ and the GMM-assigned cluster labels.  
For each sample $i$, let $a(i)$ denote the mean distance to other samples within the same GMM cluster, and $b(i)$ the minimum mean distance to samples in the nearest cluster. The Silhouette value is defined as:
\begin{equation}
\psi(i) = \frac{b(i) - a(i)}{\max\{a(i), b(i)\}}.
\end{equation}
The overall clustering score is the average over all $N$ samples:
\begin{equation}
\label{eq:silhouette}
\psi = \frac{1}{N} \sum_{i=1}^{N} s(i).
\end{equation}
We accept the clustering result only if $\psi > 0.9$, ensuring that the benign and suspicious sets derived from the GMM are well separated in the PCA space. The threshold of 0.9 is chosen to impose a high-confidence separation criterion, as values above this level indicate that clusters are both compact and well distinguished from one another~\cite{shahapure2020cluster}.

\subsection{Data-Driven Blurring}

Let \(I \in \mathbb{R}^{C \times H \times W}\) be an input image with \(C\) channels, height \(H\), and width \(W\).  
For each image marked as malicious in \(C_{S,y}\), we compute a saliency map  
\(S \in \mathbb{R}^{H \times W}\) that highlights regions likely to contain trigger patterns. Here, \(h \in \{1, \ldots, H\}\) and \(w \in \{1, \ldots, W\}\) denote the pixel coordinates.  
The saliency map is obtained from the first principal component of a pixel-space PCA trained on malicious samples, as it captures the direction of maximum shared variance among these samples corresponding to the consistent trigger pattern embedded across them.  
This component is reshaped into image space, combined across channels using the Euclidean norm, and then normalized to the range \([0,1]\):  

\begin{equation}
\label{eq:saliency_pca}
S(h,w) = 
\frac{
\sqrt{\sum_{c=1}^{C} \big(\text{PC}_1(c,h,w)\big)^2} - \min(S)
}{
\max(S) - \min(S) + \epsilon
},
\end{equation}

where \(\text{PC}_1\) is the first principal component in image space, and  
\(\epsilon > 0\) is a small constant for stability.  

Hereafter a binary mask \(M_{\tau} \in \{0,1\}^{H \times W}\) is then created by thresholding the saliency map:  

\begin{equation}
\label{eq:mask}
M_{\tau}(u,v) =
\begin{cases}
1, & \text{if } S(u,v) \geq \tau,\\
0, & \text{otherwise.}
\end{cases}
\end{equation}

Next, we identify which images should be blurred.  
Let \(z_p \in \mathbb{R}^2\) be the PCA embedding of image \(p\) in $C_{B,y}$, and let \(\mu_{\mathrm{mal}}\) be the mean embedding of malicious samples.  
The Euclidean distance between \(z_p\) and the malicious center is:  

\begin{equation}
\label{eq:distance}
d_p = \lVert z_p - \mu_{\mathrm{mal}} \rVert_2
\end{equation}

The top \(K\) images with the smallest distances are flagged as suspicious.  
Only this subset, denoted by \(\mathcal{S'}\), is processed:  

\begin{equation}
\label{eq:suspecious}
S' = \max\{0, \lfloor K * N \rfloor \}
\end{equation}

For each suspicious image, the blur strength depends on its distance to the malicious center.  
Distances are converted into a closeness score:  

\begin{equation}
\label{eq:disblur}
\delta_p = 1 - \frac{d_p}{\max_{j \in \mathcal{S'}} d_j}, \quad \delta_p \in [0,1],
\end{equation}

which is then mapped to a Gaussian blur level:  

\begin{equation}
\label{eq:gaussian}
\sigma_p = \sigma_{\min} + \delta_p \cdot (\sigma_{\max} - \sigma_{\min}),
\end{equation}

with \(\sigma_{\min} \geq 0\) and \(\sigma_{\max} > \sigma_{\min}\).  

The final defended image is obtained by replacing suspicious pixels of all images in $S'$ with their blurred versions:  

\begin{equation}
\label{eq:blur}
\begin{split}
I'(u,v,c) &= \big(1 - M_{\tau}(u,v)\big)\, I(u,v,c) \\
&\quad+ M_{\tau}(u,v)\, \big[G_{\sigma_p} \star I\big](u,v,c),
\end{split}
\end{equation}

for channels \(c = 1,\dots,C\), where \(G_{\sigma_p}\) is a Gaussian kernel with standard deviation \(\sigma_p\).  
Images not in \(\mathcal{S'}\) remain unchanged.  

This formulation ensures that malicious samples are excluded, as well as the removal or weakening of potential triggers. Consequently, retaining essential semantic information which provides maximal protection against backdoor contamination while preserving model utility on normal tasks.

\section{Poisoned Data Filtering in Federated Learning}
\label{FL-PBM}

In this section, we present the operational flow of our pre-training backdoor mitigation approach in the FL environment, along with the algorithms representing each step of the process.

\begin{figure*}[!t]
    \centering
    \includegraphics[width=0.99\textwidth]{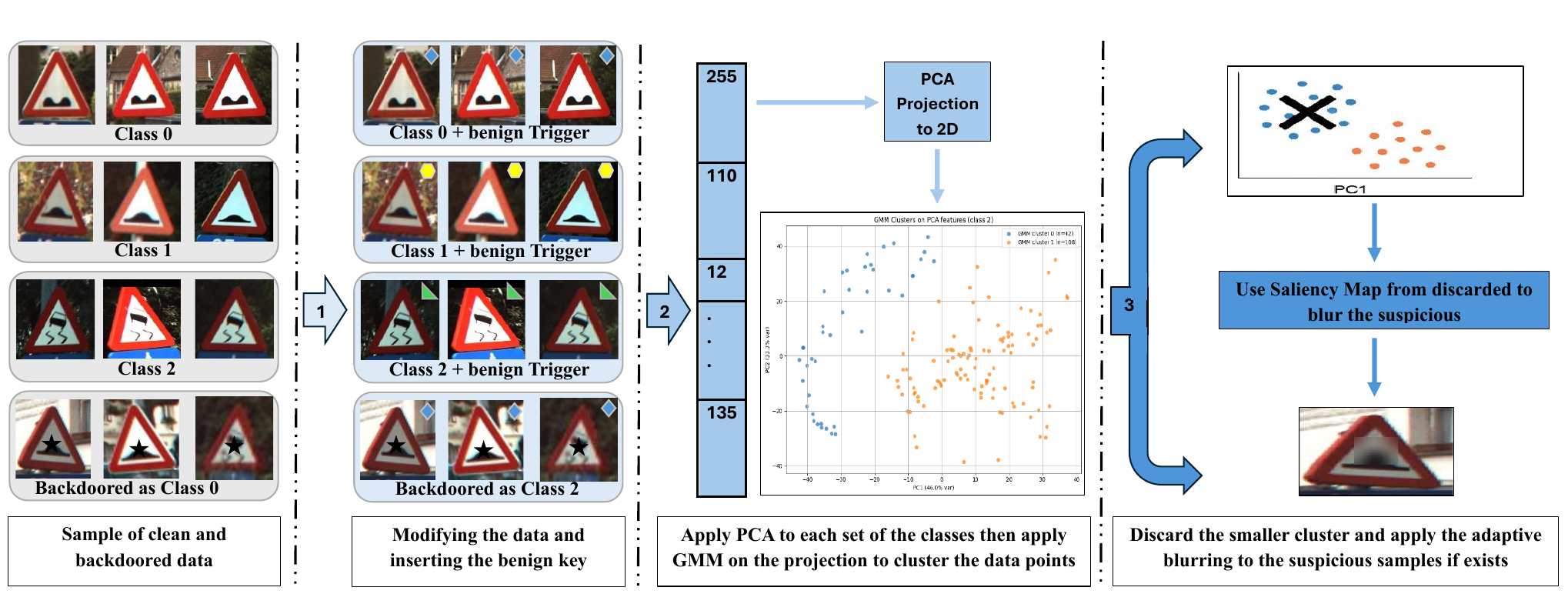}
    \caption{Overview of the proposed client-side preprocessing pipeline for backdoor mitigation in federated learning. The main components include dataset preprocessing, benign trigger augmentation, dimensionality reduction, clustering, and selective sanitization.}
    \label{fig:sys_arch}
\end{figure*}

Fig.~\ref{fig:sys_arch} shows the proposed architecture, which adds a client-side preprocessing pipeline to reduce backdoor attacks in FL. All preprocessing steps are executed locally on each client under server orchestration, ensuring that data privacy is preserved while the server coordinates the backdoor mitigation process before training begins. The process starts with the client’s local dataset, which can contain both clean and poisoned samples (marked with a black star). To build a consistent baseline and make classes easier to separate, a benign trigger is added to all class samples, creating augmented references (Step~1). Next, Principal Component Analysis (PCA) is used to project the data into a lower-dimensional space, keeping the main variation while reducing noise. In this space, Gaussian Mixture Models (GMM) are applied to group the data. Smaller or isolated groups are marked as suspicious. Samples from these groups are treated in two ways: clearly abnormal samples are removed, while moderately suspicious ones are cleaned further (Step~2). For the removed samples, saliency maps are created to locate possible trigger regions. Adaptive blurring is then applied to a subset of the kept samples that are very close to the center of the removed group. This step helps weaken potential triggers while keeping the main image structure (Step~3). The final cleaned dataset, without highly suspicious samples and with weakened triggers, is then used for local training, reducing the risk of backdoor attacks while maintaining model accuracy on normal data.

\begin{algorithm}
  \caption{Client-Side Preprocessing and Local Training}
  \label{alg:clustering}
  \begin{algorithmic}[1]
  \small
    \Require Local dataset $\mathcal{D}_i$ for client $i$, number of epochs $E$, Initial model $\mathcal{G}_i^t$ 
    \Ensure Updated local model $\mathcal{G}_i^{t+1}$
    \For{each class $c \in \mathcal{C}_i$}
        \State Insert benign trigger into class samples
        \State Apply PCA on $\mathcal{D}_i^{(c)}$ using Eq.~\ref{eq:pca}
        \State Fit GMM on PCA features using Eq.~\ref{eq:fit}
        \State Assign each sample to a cluster using Eq.~\ref{eq:assi}
        \State Identify benign ($C_{B,y}$) and suspicious ($C_{S,y}$) clusters using Eq.~\ref{eq:cluster}
        \State Approve clustering using Silhouette score (Eq.~\ref{eq:silhouette})
        \State Remove benign triggers from samples
        \State Apply Data-Driven Blurring on ($C_{B,y}, C_{S,y}$) (Alg.~\ref{alg:blurring})
    \EndFor
    \For{each epoch $e = 1$ to $E$}
        \State Train local model on preprocessed dataset $\mathcal{D}'_i$
    \EndFor
    \State Send updated local model $\mathcal{G}_i^{t+1}$ to server
  \end{algorithmic}
\end{algorithm}

Algorithm~\ref{alg:clustering} outlines the client-side preprocessing phase, which applies PCA-based dimensionality reduction, GMM clustering, and Silhouette-based validation, followed by the federated training iterations. For every class $c$ in client $i$ local dataset $\mathcal{D}_i$, a benign trigger is first inserted to help reveal potential distributional irregularities (Line~2). PCA is then applied to the class-specific data $\mathcal{D}_i^{(c)}$ to obtain low-dimensional representations (Line~3). Using these features, a two-component GMM is fitted to capture the data distribution and detect hidden data clusters (Line~4). Each sample is then assigned to a cluster using the MAP rule (Line~5), after which the benign and suspicious clusters are identified (Line~6). The clustering result is then validated using the Silhouette Coefficient; only clusters achieving a score $S>0.9$ are accepted as well-separated (Line~7). After validation, benign triggers are removed to restore the data to its clean state (Line~8), and a Data-Driven Blurring process is applied to the identified clusters $(C_{B,y}, C_{S,y})$ to mitigate potential backdoor effects (Line~9). Once preprocessing is complete, the model training is performed on the client's processed dataset $\mathcal{D}'_i$ over $E$ epochs (Lines~11–13). After completing local training, the updated model $\mathcal{G}_i^{t+1}$ is sent to the server (Line~14) for aggregation.

\begin{algorithm}
    \caption{Data-Driven Blurring of Suspicious Samples}
    \label{alg:blurring}
    \begin{algorithmic}[1]
    \small
    \Require Benign cluster $C_{B,y}$, malicious cluster $C_{S,y}$, parameters $k$
    \Ensure Blurred suspicious samples within $C_{B,y}$
    \State Compute saliency map $S$ from $C_{S,y}$ using Eq.~\ref{eq:saliency_pca}
    \State Generate binary mask $M_\tau$ using Eq.~\ref{eq:mask}
    \State $S' \leftarrow$ Select top $k$ nearest samples from $C_{B,y}$ using Eq.~\ref{eq:suspecious}
    \For{each suspicious sample $I_i \in S'$}
        \State Compute distance from malicious set $C_{S,y}$ using Eq.~\ref{eq:distance}
        \State Assign blur strength using Eq.~\ref{eq:disblur}
        \State Compute Gaussian kernel $G_{\sigma_i}$ using Eq.~\ref{eq:gaussian}
        \State Apply blur to suspicious regions of $I_i$ using Eq.~\ref{eq:blur}
    \EndFor
    \end{algorithmic}
\end{algorithm}

Algorithm~\ref{alg:blurring} outlines the procedure for applying data-driven
blurring to suspicious samples. First, a normalized saliency map $S$ is computed from the first PCA component from each sample of the suspicious cluster $C_{S,y}$ to highlight potential trigger regions (Line~1). The saliency map is then converted into a binary mask $M_\tau$ by thresholding at level $\tau$ (Line~2). Next, the top $k$ nearest suspicious samples $S'$ are selected from the benign cluster $C_{B,y}$ using, based on their proximity to the malicious distribution (Line~3). For each suspicious sample $I_i \in S'$, the distance $d_i$ between the sample’s embedding $z_i$ and the malicious cluster center $\mu_{\text{mal}}$ is calculated (Line~5). Each sample is then assigned a blur strength according to its normalized distance from the malicious center (Line~6). A Gaussian kernel $G_{\sigma_i}$ with standard deviation $\sigma_i$ is constructed to define the blur level (Line~7). Finally, the blur is applied selectively to the suspicious regions of the image, as indicated by the mask $M_\tau$, producing a sanitized version $I'_i$ of $I_i$ (Line~8). This process reduces or removes possible triggers while preserving the overall structure of the data.

\section{Experimental Evaluation}
\label{exp}
Hereafter, we evaluate the proposed \textit{FL-PBM} to demonstrate its effectiveness against different types of backdoor attacks. We first describe the experimental setup, including datasets, poisoning configurations, and defense baselines, followed by quantitative and visual comparisons across \emph{IID} and \emph{non-IID} settings. The analysis highlights how \textit{FL-PBM} mitigates backdoor effects while preserving model utility across both balanced and highly skewed data distributions.

\subsection{Datasets and Attack Setup}
We evaluate our approach on two traffic-sign benchmarks, the German Traffic Sign Recognition Benchmark (GTSRB)~\cite{stallkamp2011german} and the Belgian Traffic Sign Classification dataset (BTSC)~\cite{mathias2013traffic}. From each benchmark, we derive a 10-class subset (\emph{GTSRB-10}, \emph{BTSC-10}) by selecting the ten classes with the largest number of samples. For the attack setting, a triangular trigger is placed in the image corner with a side length of $5$ pixels and an opacity of $0.5$. We evaluate both \emph{one-to-n} and \emph{n-to-n} attack paradigms. For a malicious client, a fraction of the source-class samples is poisoned; this fraction is varied between $0.30$ and $0.70$ to explore weak-to-strong poisoning scenarios. At the client-selection level, malicious participants constitute $30\%$ of the selected clients at every communication round. 

Since our approach is applied at the data level, it is essential to consider various data distributions in the experimental setup. We run experiments under two data-partitioning scenarios to reflect realistic federated and centralized deployments:

    \paragraph{Non-IID} Clients exhibit heterogeneity across classs, where each client holds between $3$ and $6$ classes, with $100$--$200$ images per class. From a population of $100$ clients we sample $40$ clients per round to participate in training for the GTSRB dataset and each client holds between $3$ and $4$ classes, with $30$--$60$ images per class. From a population of $100$ clients we sample $20$ clients per round to participate in training for the BTSC dataset; malicious clients among those selected follow the poisoning procedure described above. This configuration models practical class imbalance and distribution skew encountered in cross-device FL environments. On the other hand, for the centralized approach, a class size imbalance is applied.

    \paragraph{IID} For centralized (non-federated) experiments, we enforce class balance via oversampling so that all classes have equal size. For federated IID experiments, each client contains all $10$ classes with $150$ images per class for the GTSRB dataset and $30$ samples per class for the BTSC, thereby isolating the impact of statistical heterogeneity from other system-level factors. All other experimental settings (trigger positions, poisoning rates, and fraction of malicious clients) are kept identical between IID and non-IID runs to ensure fair comparison.

\begin{figure*}[!t]
  \centering
  \subfloat[\scriptsize GTSRB IID - (1$\rightarrow$1) Attack]{\includegraphics[width=0.25\textwidth,height=3cm]{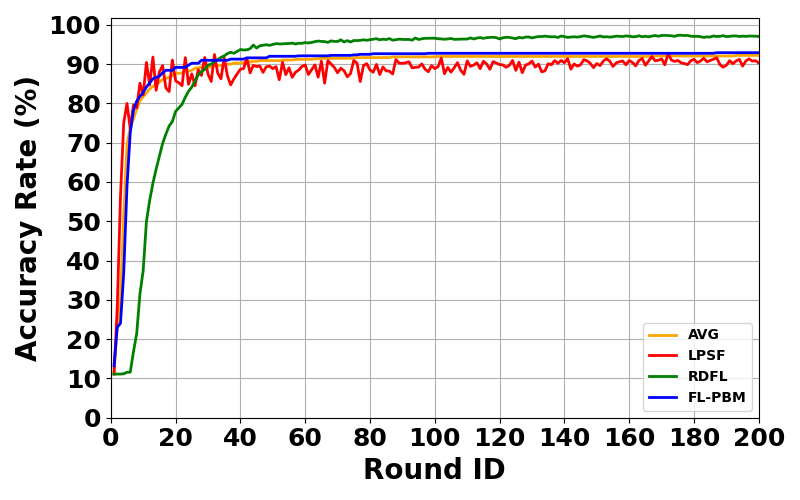}}
  \subfloat[\scriptsize GTSRB IID - (N$\rightarrow$1) Attack]{\includegraphics[width=0.25\textwidth,height=3cm]{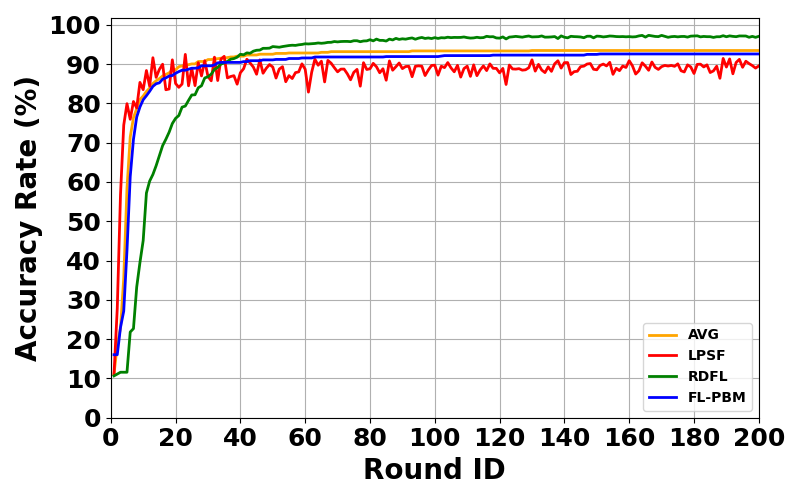}}
  \subfloat[\scriptsize GTSRB non-IID - (1$\rightarrow$1) Attack]{\includegraphics[width=0.25\textwidth,height=3cm]{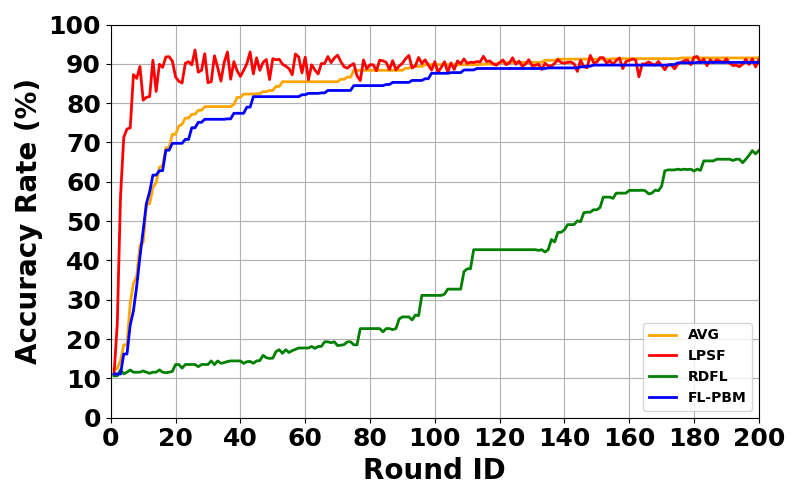}}
  \subfloat[\scriptsize GTSRB non-IID - (N$\rightarrow$1) Attack]{\includegraphics[width=0.25\textwidth,height=3cm]{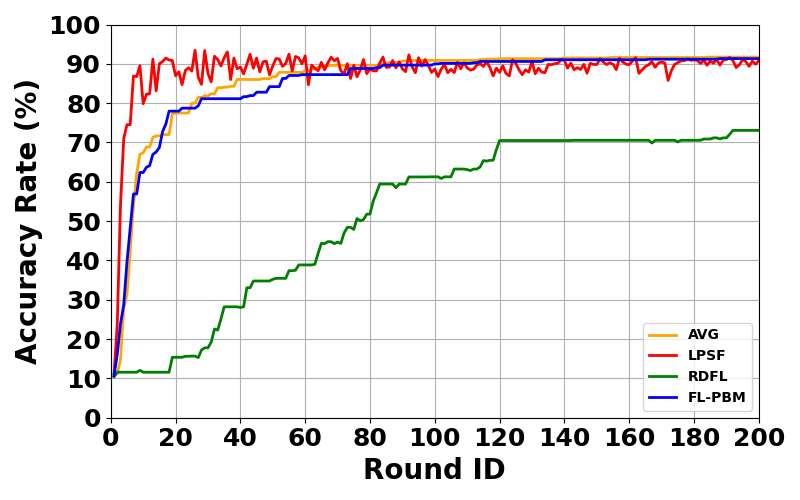}}\\
  \subfloat[\scriptsize BTSC IID - (1$\rightarrow$1) Attack]{\includegraphics[width=0.25\textwidth,height=3cm]{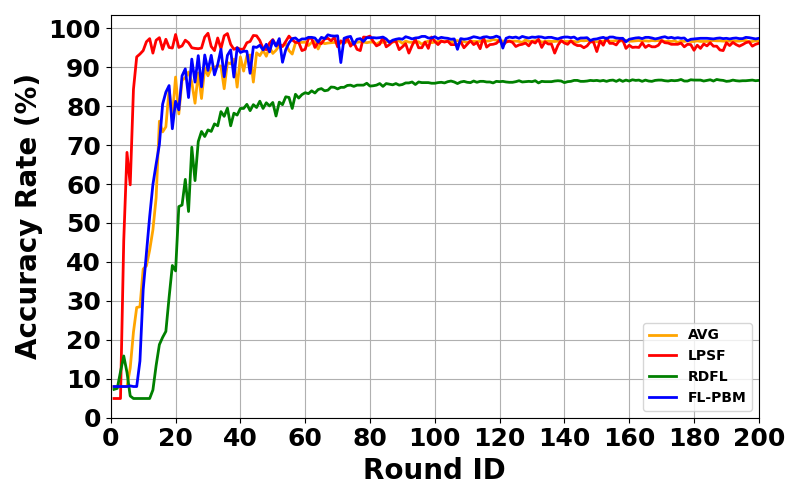}}
  \subfloat[\scriptsize BTSC IID - (N$\rightarrow$1) Attack]{\includegraphics[width=0.25\textwidth,height=3cm]{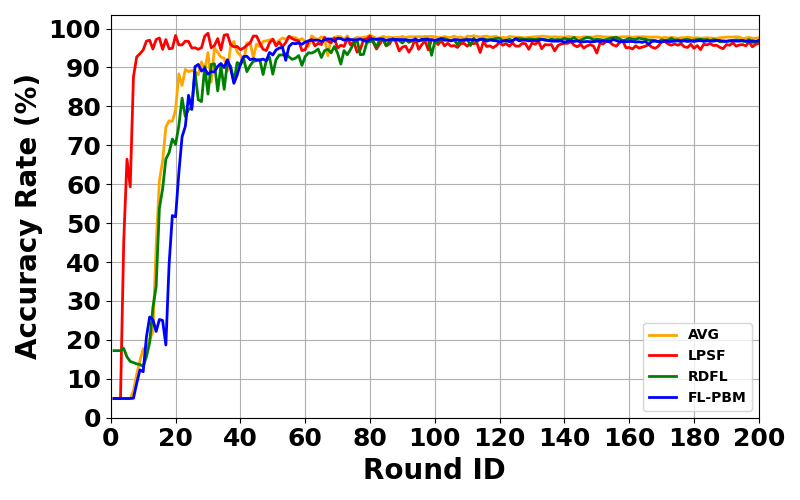}}
  \subfloat[\scriptsize BTSC non-IID - (1$\rightarrow$1) Attack]{\includegraphics[width=0.25\textwidth,height=3cm]{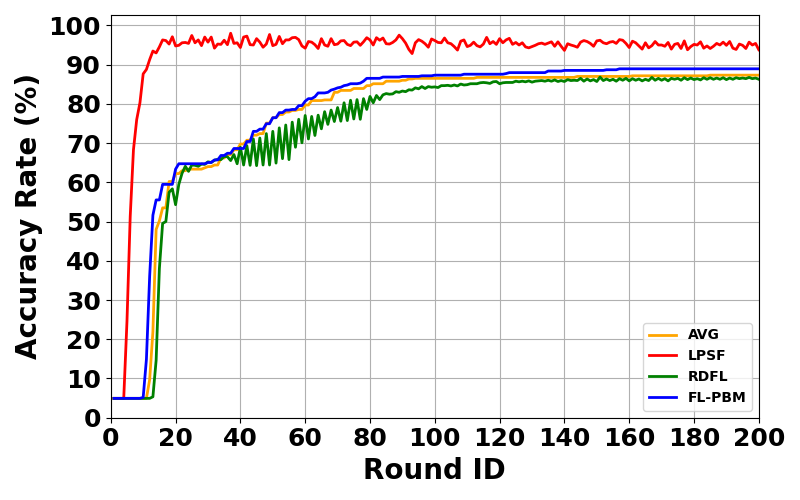}}
  \subfloat[\scriptsize BTSC non-IID - (N$\rightarrow$1) Attack]{\includegraphics[width=0.25\textwidth,height=3cm]{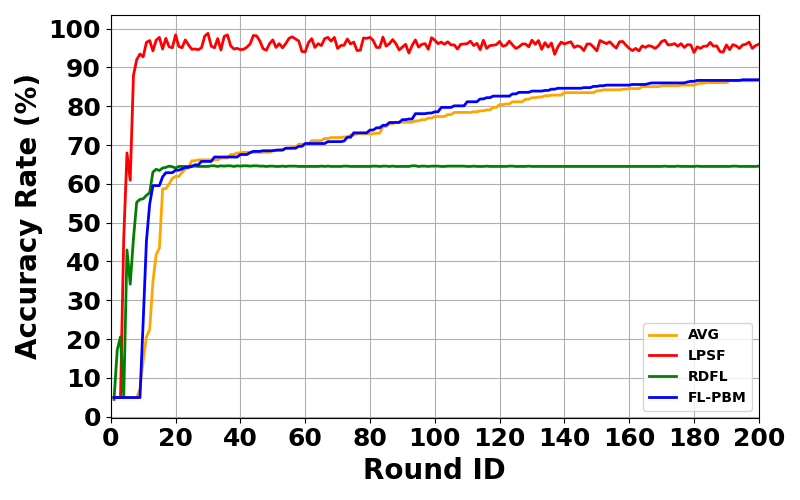}}
  \caption{Model accuracy of different approaches over the normal task across datasets and distributions over 200 rounds. The color mapping for defense strategies is AVG (orange), LPSF (red), RDFL (green), FL-PBM (blue)}
  \label{fig:acc_grid}
\end{figure*}

\begin{figure*}[!t]
  \centering
  \subfloat[\scriptsize GTSRB IID - (1$\rightarrow$1) Attack]{\includegraphics[width=0.25\textwidth,height=3.2cm]{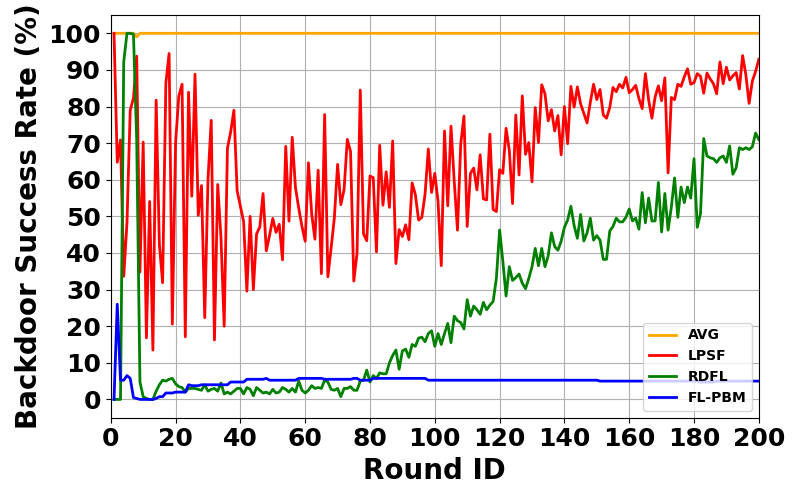}}
  \subfloat[\scriptsize GTSRB IID - (N$\rightarrow$1) Attack]{\includegraphics[width=0.25\textwidth,height=3.2cm]{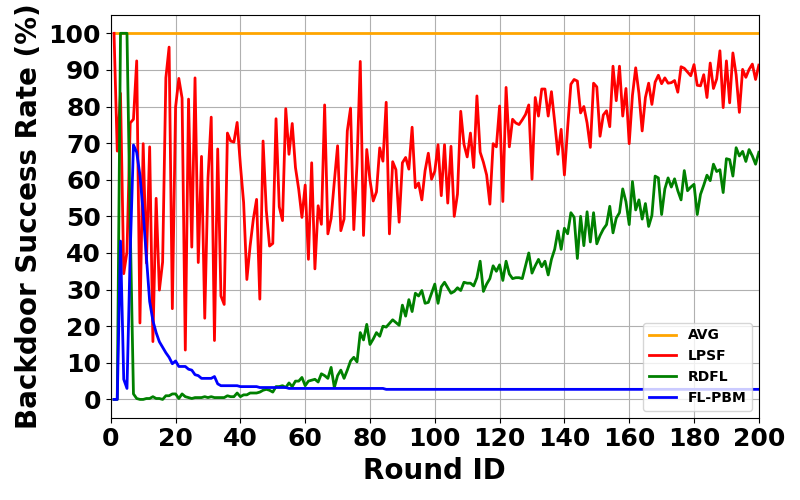}}
  \subfloat[\scriptsize GTSRB non-IID - (1$\rightarrow$1) Attack]{\includegraphics[width=0.25\textwidth,height=3.2cm]{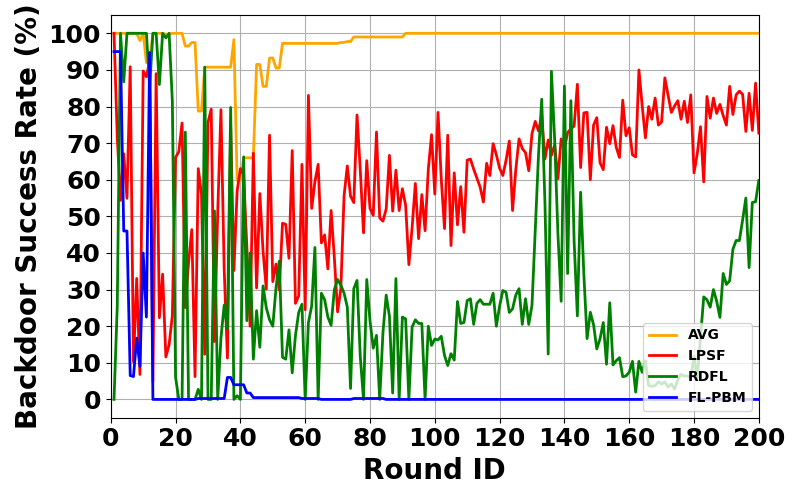}}
  \subfloat[\scriptsize GTSRB non-IID - (N$\rightarrow$1) Attack]{\includegraphics[width=0.25\textwidth,height=3.2cm]{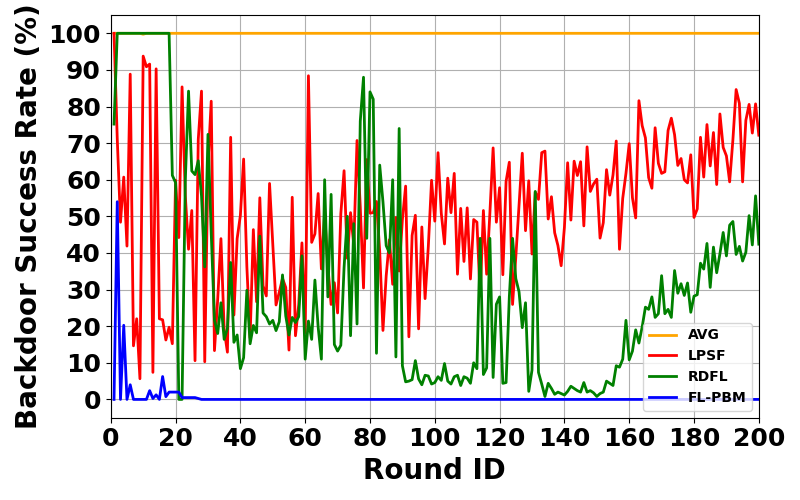}}\\
  \subfloat[\scriptsize BTSC IID - (1$\rightarrow$1) Attack]{\includegraphics[width=0.25\textwidth,height=3.2cm]{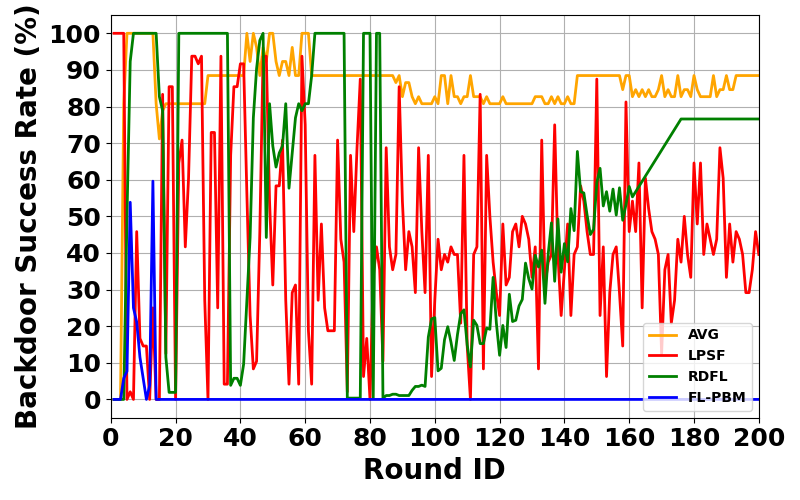}}
  \subfloat[\scriptsize BTSC IID - (N$\rightarrow$1) Attack]{\includegraphics[width=0.25\textwidth,height=3.2cm]{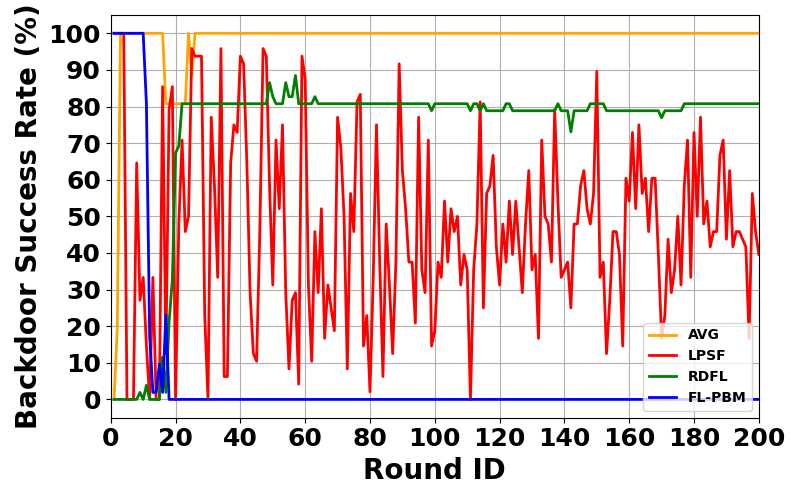}}
  \subfloat[\scriptsize BTSC non-IID - (1$\rightarrow$1) Attack]{\includegraphics[width=0.25\textwidth,height=3.2cm]{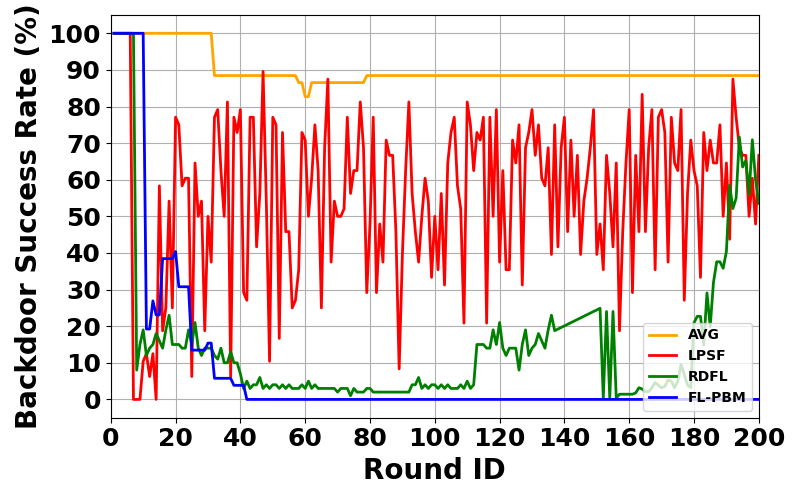}}
  \subfloat[\scriptsize BTSC non-IID - (N$\rightarrow$1) Attack]{\includegraphics[width=0.25\textwidth,height=3.2cm]{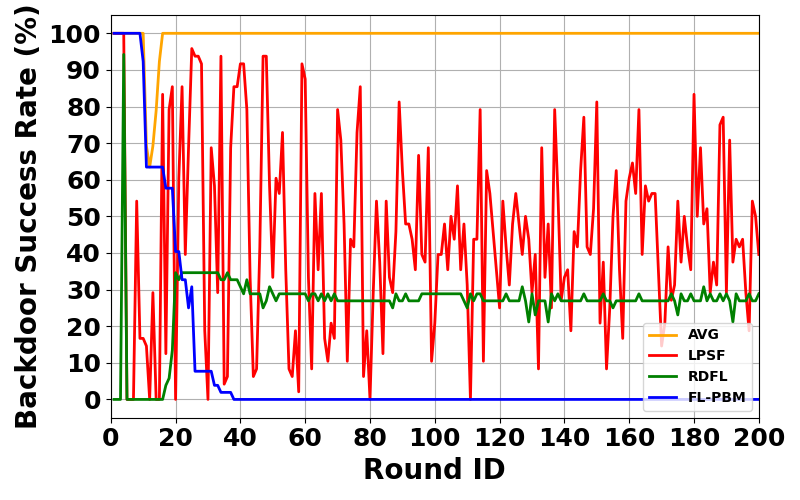}}
  \caption{Model accuracy of different approaches over the malicious task across datasets and distributions for 200 rounds. The color mapping for defense strategies is AVG (orange), LPSF (red), RDFL (green), FL-PBM (blue)}
  \label{fig:bacc_grid}
\end{figure*}

\subsection{Comparison Benchmarks and Metrics}
For all experiments, we adopt the LeNet-5 architecture, modified to accommodate RGB input and $10$ output classes. Training is performed using stochastic gradient descent (SGD) with a learning rate of $0.01$, batch size of $64$, and $5$ local epochs per client update. The global training process spans $200$ communication rounds. To assess attack effectiveness, each run includes $400$ backdoored test samples for the GTSRB dataset and $200$ samples for the BTSC dataset embedded with the triangular trigger.

We benchmark three aggregation strategies representing baseline, centralized, and federated defense paradigms:
\begin{itemize}
  \item \textbf{FedAvg (baseline):} the standard federated averaging algorithm, serving as a reference to assess the impact of backdoor attacks and the effectiveness of defense mechanisms~\cite{mcmahan2017communication}.
  \item \textbf{LPSF:} a centralized defense method that identifies and filters suspicious client updates before aggregation, thereby reducing the influence of poisoned contributions~\cite{kaviani2023defense}.
  \item \textbf{RDFL:} a federated defense algorithm that dynamically adapts aggregation weights to downscale malicious updates during the model aggregation phase~\cite{wang2023adaptive}. \\
\end{itemize}

In our experiments, model performance is quantified using two main metrics:

\begin{itemize}
  \item \textbf{Clean Accuracy Rate (CAR):} classification accuracy on clean test samples, measuring the model’s utility and convergence behavior as well as the impact of the proposed solution on the convergence of the global model.
  \item \textbf{Attack Success Rate (ASR):} the proportion of triggered inputs misclassified into the target class, capturing the effectiveness of the proposed approach in mitigating the backdoor attack effect on the global model.
\end{itemize}
These metrics provide a balanced evaluation of both model utility under normal operation and resilience against adversarial manipulation.

Results are presented for both non-IID and IID scenarios, where each outcome is depicted as curves of CAR and ASR over communication rounds. Figures~\ref{fig:acc_grid} and~\ref{fig:bacc_grid} show results for the normal and malicious tasks, respectively, across datasets, distributions, and attack types. Different defense strategies are distinguished by color, following the mapping AVG (orange), LPSF (red), RDFL (green), and FL-PBM (blue).

\subsection{Results on IID Setting}
\label{sec:results-iid}

We first examine the IID results, corresponding to subfigures (a), (b), (e), and (f) in Figures~\ref{fig:acc_grid} and~\ref{fig:bacc_grid}. Each figure overlays clean accuracy rate (CAR) and attack success rate (ASR) across 200 communication rounds, allowing us to track defense evolution from initialization to convergence, as well as the interplay between attack intensity and mitigation strategies.

\noindent\textbf{(a) GTSRB IID, $(1\!\rightarrow\!1)$ Attack.}
In the first 20 rounds, \textit{FedAvg} is rapidly compromised, with ASR exceeding 90\% and CAR below 40\%, reflecting the dominance of backdoor updates. \textit{LPSF} mitigates the attack partially by pruning weakly active connections, reducing ASR to approximately 70\%, but also removes some informative features, keeping CAR below 50\%. \textit{RDFL} apply training via adaptive aggregation, achieving CAR near 55\% and ASR around 40\%, showing a trade-off between robustness and accuracy. \textit{FL-PBM} demonstrates early resilience, filtering and blurring suspicious samples before local training, which maintains CAR above 55\% and lowers ASR below 10\%.  
By 100 rounds, \textit{FedAvg} remains highly vulnerable CAR $\approx$ 91\%, ASR above 90\%, whereas \textit{LPSF} lowers ASR to around 60\% but CAR remains under 90\%. \textit{RDFL} improves both metrics with CAR $\approx$ 94\% and ASR $\approx$ 20\%, while \textit{FL-PBM} continues to outperform, maintaining CAR above 90\% and ASR under 10\%. By 200 rounds, \textit{FL-PBM} consistently achieves CAR above 93\% and ASR close to 5\%, within normal model misclassification rates, confirming the superiority of proactive client-side mitigation over reactive aggregation-based defenses.

\vspace{0.4em}
\noindent\textbf{(b) GTSRB IID, $(N\!\rightarrow\!1)$ Attack.}
During early rounds, \textit{FedAvg} is overwhelmed with ASR above 90\%, CAR $\approx$ 88\%, while \textit{LPSF} reduces ASR to 70\%, while keeping CAR near 87\%. \textit{RDFL} shows early stability with CAR $\approx$ 77\%, ASR $\approx$ 0\%, and \textit{FL-PBM} prevents most backdoor effects with CAR above 87\% and ASR below 15\%. At 100 rounds, \textit{FedAvg} remains highly susceptible at CAR $\approx$ 93\%, ASR above 90\%, \textit{LPSF} lowers ASR to $\sim$60\%, and \textit{RDFL} reaches CAR near 96\% with ASR $\approx$ 30\%. \textit{FL-PBM} sustains CAR above 90\% and ASR under 4\%, confirming its effectiveness against multi-source attacks. By 200 rounds, only \textit{FL-PBM} consistently maintains CAR above 93\% and ASR below 5\%.

\vspace{0.4em}
\noindent\textbf{(e) BTSC IID, $(1\!\rightarrow\!1)$ Attack.}
Initially, \textit{FedAvg} is highly vulnerable with ASR greater than 80\%, CAR lower than 80\% due to the quick dominance of poisoned updates in the smaller dataset. \textit{LPSF} reduces ASR to $\sim$60\%, but excessive pruning keep CAR below 50\%. \textit{RDFL} achieving high CAR $\approx$ 97\%, and moderate ASR $\approx$ 40\%, whereas \textit{FL-PBM} benefits from early filtering, achieving CAR above 80\% and ASR near 0\%. By 100 rounds, \textit{FL-PBM} maintains CAR above 97\% with ASR $\approx$ 0\%, clearly outperforming baselines. By 200 rounds, \textit{FL-PBM} consistently shows strong defense with stable ASR at $\approx$ 0\%, highlighting the advantage of proactive client-side mitigation when per-class data is limited.

\vspace{0.4em}
\noindent\textbf{(f) BTSC IID, $(N\!\rightarrow\!1)$ Attack.}
Early on, \textit{FedAvg} collapses with ASR above 80\%, CAR $\approx$ 80\%. \textit{LPSF} reduces ASR to 68\% while achieving high CAR above 95\%, \textit{RDFL} achieves CAR $\approx$ 70\%, with moderate ASR $\approx$ 45\%, and \textit{FL-PBM} achieves early robustness with CAR above 60\%, while reducing ASR to around 0\%. By 100 rounds, \textit{FL-PBM} maintains CAR above 95\% with negligible ASR. By 200 rounds, only \textit{FL-PBM} consistently sustains CAR above 96\% and ASR near 0\%, confirming the effectiveness of PCA+GMM clustering combined with blurring transformations for neutralizing one trigger backdoor attack.

\subsection{Results on Non-IID Setting}
\label{sec:results-noniid}

The non-IID scenarios, shown in subfigures (c), (d), (g), and (h) of Figures~\ref{fig:acc_grid} and~\ref{fig:bacc_grid}, pose greater challenges due to class imbalance and heterogeneous client distributions. Tracking CAR and ASR across rounds reveals the resilience of different defense mechanisms under these conditions.

\noindent\textbf{(c) GTSRB non-IID, $(1\!\rightarrow\!1)$ Attack.}
Early rounds show \textit{FedAvg} instability with ASR above 85\%, CAR nearly 70\%, \textit{LPSF} reduces ASR to 65\% while increasing CAR to 89\%, whereas \textit{RDFL} exhibits low stability with CAR $\approx$ 23\%, ASR $\approx$ 40\%. \textit{FL-PBM} achieves early robustness, maintaining CAR above 70\% and ASR nearly 0\%. At 100 rounds, \textit{FedAvg} remains fully compromised with CAR $\approx$ 90\%, ASR $\approx$ 100\%, \textit{LPSF} lowers ASR to 55\%, \textit{RDFL} records CAR around 32\%, while \textit{FL-PBM} sustains CAR above 85\% and ASR near 0\%. By 200 rounds, \textit{FL-PBM} achieves CAR $\approx$ 90\% and ASR $\approx$ 0\%, demonstrating effectiveness under heterogeneous distributions.

\vspace{0.4em}
\noindent\textbf{(d) GTSRB non-IID, $(N\!\rightarrow\!1)$ Attack.}
In early rounds, \textit{FedAvg} collapses ASR above 90\%, CAR less than 30\%, \textit{LPSF} reduces ASR to around 70\% with CAR lower than 50\%, \textit{RDFL} stabilizes moderately with CAR $\approx$ 55\%, ASR $\approx$ 45\%, and \textit{FL-PBM} reaches CAR above 60\% with ASR below 25\%. At 100 rounds, \textit{FL-PBM} sustains CAR above 78\% and ASR under 15\%, outperforming baselines. By 200 rounds, \textit{FL-PBM} achieves CAR above 86\% and ASR near 0\%, confirming effective prevention of poisoned samples propagation to the training process.

\vspace{0.4em}
\noindent\textbf{(g) BTSC non-IID, $(1\!\rightarrow\!1)$ Attack.}
Early rounds show \textit{FedAvg} vulnerable with ASR above 90\%, CAR around 60\%, \textit{LPSF} reduces ASR to 80\% while CAR nearly 95\%, \textit{RDFL} stabilizes the model with CAR nearly 55\%, ASR less than 20\%, and \textit{FL-PBM} reduces ASR below 40\% while CAR remains above 60\%. By 100 rounds, \textit{FL-PBM} maintains CAR above 85\% and ASR near 0\%, and by 200 rounds, CAR reaches 89\% with ASR consistently at 0\%.

\vspace{0.4em}
\noindent\textbf{(h) BTSC non-IID, $(N\!\rightarrow\!1)$ Attack.}
\textit{FedAvg} get compromised rapidly with ASR above 90\%, CAR around 62\%, \textit{LPSF} reduces ASR to 60\%, \textit{RDFL} stabilizes moderately at CAR $\approx$ 65\%, ASR $\approx$ 35\%, and \textit{FL-PBM} maintains CAR above 63\% with ASR below 10\%. By 200 rounds, \textit{FL-PBM} consistently outperforms all baselines with CAR around 86\%, ASR near 0\%, demonstrating that \textit{FL-PBM} effectively neutralize multi-trigger backdoor attacks under heterogeneous settings.

\subsection{Discussion}
Across both IID and non-IID scenarios, the comparative analysis highlights clear differences in the behavior of the evaluated approaches. \textit{FedAvg}, which lacks any defensive mechanism, is consistently the most vulnerable, where its clean accuracy rate (CAR) remains modest while its attack success rate (ASR) stays persistently high throughout training. \textit{LPSF} provides partial robustness by filtering suspicious updates, but its centralized filtering often discards some benign input feature connections, which slows convergence and limits final accuracy. \textit{RDFL} performs better by dynamically adjusting aggregation weights, achieving a more balanced trade-off between CAR and ASR, yet it cannot fully suppress backdoor effects, especially under stronger $(N \rightarrow 1)$ attacks and in non-IID settings due to its reliance on the model's similarity, which can be affected by the data distribution as well as the model convergence. 

In contrast, \textit{FL-PBM} consistently demonstrates superior resilience across all datasets, attack types, and data distributions. Its proactive client-side filtering, which combines PCA-based feature reduction with Gaussian Mixture Model clustering and targeted blurring of suspicious samples, prevents and limits poisoned data from influencing local training. This explains why \textit{FL-PBM} achieves high CAR (often above 80\%) while reducing ASR to near-zero levels, even in the most challenging non-IID and multi-class poisoning scenarios. These results confirm that addressing backdoor threats at the data level before aggregation is more effective than relying solely on weighting aggregation or input feature pruning defences.

\section{Conclusion}
\label{conc}
In this work, we proposed \textit{FL-PBM}, a novel backdoor mitigation for FL. Unlike reactive aggregation-based defenses, \textit{FL-PBM} operates proactively at the client side by identifying and neutralizing poisoned data before it can influence the global model. By combining dimensionality reduction through PCA, unsupervised clustering with Gaussian Mixture Models, and targeted image blurring, our method effectively suppresses diverse backdoor attacks while preserving model integrity and performance. Extensive experiments on both IID and non-IID settings with GTSRB and BTSC datasets show that \textit{FedAvg} remains highly vulnerable, with attack success rates often remaining near 90--100\% even after convergence. Centralized mitigation \textit{LPSF} improves robustness but at the cost of model learning, with ASR commonly reduced only to 50--70\% while CAR drops by 15\% to 40\% because benign information is lost during filtering. \textit{RDFL} achieves a more balanced outcome, with ASR reduced to 29--80\% and CAR stabilizing between 65\% and 95\% depending on data distribution. However, performance degrades in non-IID settings where model similarity is harder to exploit. In contrast, \textit{FL-PBM} delivers the strongest mitigation across all scenarios, reducing ASR to 0\% to 5\% in most experiments while keeping CAR typically between 87\% and 97\%, even in the challenging non-IID cases with high data skew and 30\% malicious participation. These findings highlight the feasibility and effectiveness of preemptive data filtering as a powerful defense mechanism in FL based security mechanisms. As future directions, we plan to explore the integration of generative adversarial networks (GANs) trained on saliency maps of malicious samples to reproduce potential backdoor patterns and further stress-test the defense. Additionally, collecting metadata from the filtering process and leveraging advanced feature extraction methods such as autoencoders or pre-trained models may further enhance the robustness and generalization of \textit{FL-PBM}.

\section*{Acknowledgments}
This research was supported by the Natural Sciences and Engineering Research Council of Canada (NSERC). The authors gratefully acknowledge its financial support.

\bibliographystyle{IEEEtran}
\bibliography{references}

\vskip 0pt plus -1fil
\vspace{-33pt}
\begin{IEEEbiographynophoto}{Osama Wehbi} is a Ph.D. candidate in the Department of Computer and Software Engineering at Polytechnique Montreal, Canada. His primary research areas include cybersecurity, federated learning, and game theory.
\end{IEEEbiographynophoto}

\vskip 0pt plus -1fil
\vspace{-33pt}
\begin{IEEEbiographynophoto}{Sarhad Arisdakessian}  is a PhD candidate in the Department of Computer and Software Engineering at Polytechnique Montreal, Canada. His primary research interests lie in the fields of Federated Learning and Game Theory. 
\end{IEEEbiographynophoto}

\vskip 0pt plus -1fil
\vspace{-33pt}
\begin{IEEEbiographynophoto}{Omar Abdel Wahab} holds an assistant professor position with the Department of Computer and Software Engineering, Polytechnique Montreal, Canada. His current research activities are in the areas of cybersecurity, Internet of Things and artificial intelligence.
\end{IEEEbiographynophoto}

\vskip 0pt plus -1fil
\vspace{-33pt}
\begin{IEEEbiographynophoto}{Azzam Mourad}  is a professor of Computer Science with Khalifa University Abu Dhabi, UAE. His research interests include the domains of cloud computing, Artificial intelligence, and Cybersecurity.
\end{IEEEbiographynophoto}

\vskip 0pt plus -1fil
\vspace{-33pt}
\begin{IEEEbiographynophoto}{Hadi Otrok}  is a professor and chair of the Department of EECS at Khalifa University, Abu Dhabi, UAE. His research interests include the domains of computer and network security, crowd sensing and sourcing, ad hoc networks, and cloud security.
\end{IEEEbiographynophoto}

\vskip 0pt plus -1fil
\vspace{-33pt}
\begin{IEEEbiographynophoto}{Jamal Bentahar}  is a professor of Computer Science with Khalifa University Abu Dhabi, UAE. His research interests include reinforcement and deep learning, multi-agent systems, computational logics, and service computing
\end{IEEEbiographynophoto}

\end{document}